\documentclass[11pt,a4paper]{article}
\usepackage{acl2018}
\usepackage{times}
\usepackage{microtype}
\usepackage{latexsym}
\usepackage{url}
\usepackage{graphicx}
\usepackage{algorithm}
\usepackage{algpseudocode}
\usepackage{multirow}
\usepackage{xcolor}
\usepackage{amsmath,amssymb,mathtools}
\usepackage{bm}
\usepackage{CJKutf8}
\usepackage{comment}

\newcommand{\modelname}[1]{\textsc{#1}}

\newcommand{\mathboldface}[1]{\mbox{\boldmath $#1$}}
\newcommand{\VEC}[1]{\mathboldface{#1}}
\newcommand{\abs}[1]{\lvert #1 \rvert}
\newcommand{\C}[1]{\mathcal C_{w_t}^{\mathrm{#1}}}
\newcommand{\Cave}[1] {\frac{1}{\abs{\C{#1}}}\!\!\!\! \smashoperator[r]{\sum_{\;\;c\in\C{#1}}} \VEC{v}_c}
\newcommand{\CaveS}[1]{\frac{1}{\abs{\C{#1}}}\!\!\!\! \smashoperator[r]{\sum_{\;\;c\in\C{#1}}} \VEC{x}_c}

\AtBeginDocument{
\setlength\abovedisplayskip{7pt}
\setlength\belowdisplayskip{7pt}
\setlength\abovedisplayshortskip{\abovedisplayskip}
\setlength\belowdisplayshortskip{\belowdisplayskip}
}

\newcommand{\ja}[1]{\begin{CJK}{UTF8}{ipxm}{#1}\end{CJK}}

\aclfinalcopy 

\title{Unsupervised Learning of Style-sensitive Word Vectors}

\author{Reina\,Akama$^{*1}$, Kento\,Watanabe$^{\dag2}$, Sho\,Yokoi$^{*\ddag3}$, Sosuke\,Kobayashi$^{\mathsection 4}$, Kentaro\,Inui$^{*\ddag5}$ \\
        $^{*}$Graduate School of Information Sciences, Tohoku University \\
        $^{\dag}$National Institute of Advanced Industrial Science and Technology (AIST)\\
        $^{\mathsection}$Preferred Networks, Inc.\\
        $^{\ddag}$RIKEN Center for Advanced Intelligence Project \\
        \texttt{\{$^1$reina.a,$^3$yokoi,$^5$inui\}@ecei.tohoku.ac.jp,}\\
        \texttt{$^2$kento.watanabe@aist.go.jp,$^4$sosk@preferred.jp}}

\date{}
\begin{document}
\maketitle

\begin{abstract}
This paper presents the first study aimed at capturing stylistic similarity between words in an unsupervised manner.
We propose extending the continuous bag of words (CBOW) model~\cite{mikolov_13iclr} to learn style-sensitive word vectors using a wider context window under the assumption that the style of all the words in an utterance is consistent.
In addition, we introduce a novel task to predict lexical stylistic similarity and to create a benchmark dataset for this task. Our experiment with this dataset supports our assumption and demonstrates that the proposed extensions contribute to the acquisition of style-sensitive word embeddings.
\end{abstract}

\section{Introduction}

Analyzing and generating natural language texts requires the capturing of two important aspects of language: {\it what is said} and {\it how it is said}.
In the literature, much more attention has been paid to studies on {\it what is said}.
However, recently, capturing {\it how it is said}, such as stylistic variations, has also proven to be useful for natural language processing tasks such as classification, analysis, and generation~\cite{tacl16pav,emnlpws17niu,wang_17}.

This paper studies the stylistic variations of words in the context of the representation learning of words.
The lack of subjective or objective definitions is a major difficulty in studying style~\cite{emnlpws17xu}.
Previous attempts have been made to define a selected aspect of the notion of style (e.g., politeness)~\cite{mairesse_07,naacl15pavlick,acl16lucie,aaai16preotiuc,sennrich_16,niu_17};
however, it is not straightforward to create strict guidelines for identifying the stylistic profile of a given text.
The  systematic evaluations of style-sensitive word representations and the learning of style-sensitive word representations in a supervised manner
are hampered by this.
In addition, there is another trend of research forward controlling style-sensitive utterance generation
without defining the style dimensions~\cite{li_16, akama2017};
however, this line of research considers style to be something associated with a given specific character, i.e., a persona, and does not aim to capture the stylistic variation space.

\begin{figure}[t]
  \centering
  \includegraphics[width=0.9\hsize]{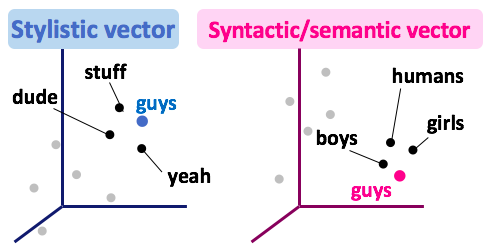}\\
  \caption{Word vector capturing stylistic and  syntactic/semantic similarity.}
  \label{fig:space}
\end{figure}

The contributions of this paper are three-fold.
(1) We propose a novel architecture that acquires style-sensitive word vectors (Figure~\ref{fig:space}) in an unsupervised manner.
(2) We construct a novel dataset for style, which consists of pairs of style-sensitive words with each pair scored according to its stylistic similarity.
(3) We demonstrate that our word vectors capture the stylistic similarity between two words successfully.
In addition, our training script and dataset are available on \url{https://jqk09a.github.io/style-sensitive-word-vectors/}.

\section{Style-sensitive Word Vector}
The key idea is to extend the continuous bag of words (CBOW)~\cite{mikolov_13iclr} by distinguishing nearby contexts and wider contexts under the assumption that a style persists throughout every single utterance in a dialog. We elaborate on it in this section.
\subsection{Notation}
Let $w_{t}$ denote the target word (token) in the corpora and $\mathcal U_t = \{w_1, \dots, w_{t-1}, w_t, w_{t+1},\dots, w_{\abs{\mathcal U_t}}\}$ denote the utterance (word sequence) including $w_t$.
Here, $w_{t+d}$ or $w_{t-d} \in \mathcal U_t$ is a context word of $w_t$ (e.g., $w_{t+1}$ is the context word next to $w_{t}$),
where $d\in\mathbb N_{>0}$ is the distance between the context words and the target word $w_t$.

For each word (token) $w$,
bold face $\VEC{v}_{w}$ and $\tilde{\VEC{v}}_{w}$ denote
the vector of $w$
and
the vector predicting the word $w$.
Let $\mathcal V$ denote the vocabulary.

\subsection{Baseline Model (\modelname{CBOW-near-ctx})}
First, we give an overview of CBOW, which is our baseline model.
CBOW predicts the target word $w_t$ given nearby context words in a window with width $\delta$:
\begin{align}
\C{near} := \left\{ w_{t\pm d} \in \mathcal U_t \mid 1\leq d \leq \delta \right\}
\end{align}
The set $\C{near}$ contains in total at most $2\delta$ words, including $\delta$ words to the left and $\delta$ words to the right of a target word.
Specifically, we train the word vectors $\tilde{\VEC{v}}_{w_t}$ and $\VEC{v}_c$ ($c\in\C{near}$) by maximizing the following prediction probability:
\begin{align}
\label{eq:w2v}
P(w_t|\C{near}) \propto \exp\biggl(\!\tilde{\VEC{v}}_{w_t} \cdot \Cave{near}\!\biggr)
\text{.}
\end{align}
The CBOW captures both semantic and syntactic word similarity through the training using nearby context words.
We refer to this form of CBOW as \modelname{CBOW-near-ctx}.
Note that, in the implementation of \citet{mikolov_13nips}, the window width $\delta$ is sampled from a uniform distribution; however, in this work, we fixed $\delta$ for simplicity.
Hereafter, throughout our experiments, we turn off the random resizing of $\delta$.

\subsection{Learning Style with Utterance-size Context Window (\modelname{CBOW-all-ctx})}
CBOW is designed to learn the semantic and syntactic aspects of words from their nearby context~\cite{mikolov_13nips}.
However, an interesting problem is determining the location where the stylistic aspects of words can be captured.  To address this problem, we start with the assumption that a style persists throughout each single utterance in a dialog, that is, the stylistic profile of a word in an utterance must be consistent with other words in the same utterance.
Based on this assumption,
we propose extending CBOW to use all the words in an utterance as context,
\begin{align}
\C{all} := \{w_{t\pm d} \in \mathcal U_t \mid 1\leq d\}
\text{,}
\end{align}
instead of only the nearby words.
Namely, we expand the context window from a fixed width to the entire utterance.
This training strategy is expected to lead to learned word vectors that are more sensitive to style rather than to other aspects.
We refer to this version as \modelname{CBOW-all-ctx}.

\subsection{Learning the Style and Syntactic/Semantic Separately}
\label{sec:proposed-method}
To learn the stylistic aspect more exclusively, we further extended the learning strategy.

\subsubsection*{Distant-context Model (\modelname{CBOW-dist-ctx})}
First, remember that using nearby context is effective for learning word vectors that capture semantic and syntactic similarities.
However, this means that using the nearby context can lead the word vectors to capture some aspects other than style.
Therefore, as the first extension, we propose excluding the \textit{nearby} context $\C{near}$ from \textit{all} the context $\C{all}$.
In other words, we use the \textit{distant} context words only:
\begin{align}
\!\C{dist} := \C{all}\setminus\C{near}
= \left\{ w_{t\pm d} \in \mathcal U_t \mid \delta < d \right\}\!\text{.}\!
\end{align}
We expect that training with this type of context will lead to word vectors containing the style-sensitive information only.
We refer to this method as \modelname{CBOW-dist-ctx}.

\subsubsection*{Separate Subspace Model (\modelname{CBOW-sep-ctx})}
\begin{figure}[t]
  \centering
  \includegraphics[width=\hsize]{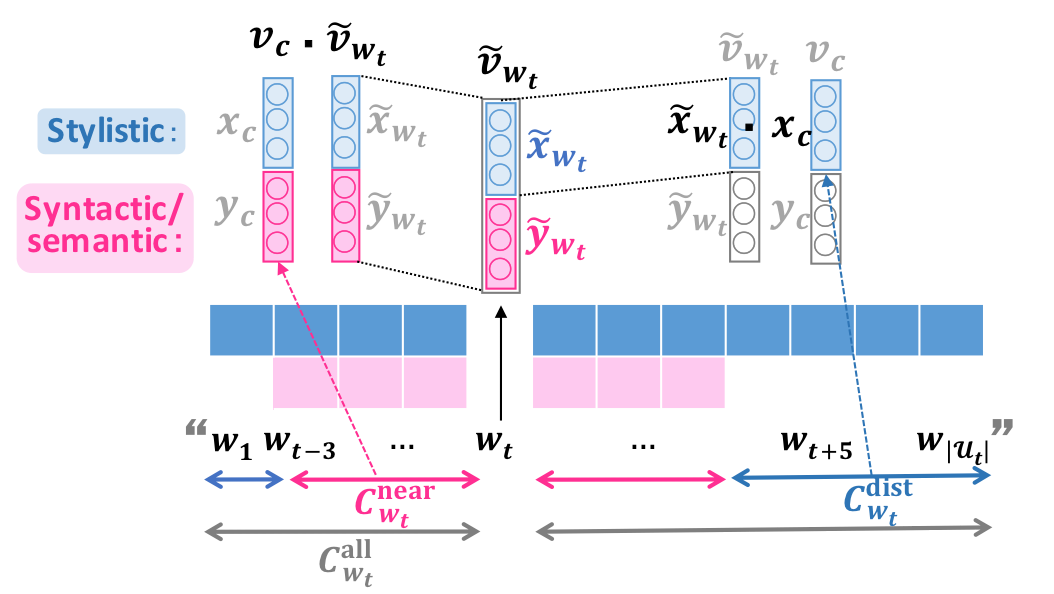}\\
  \caption{The architecture of \modelname{CBOW-sep-ctx}.}
  \label{fig:method}
\end{figure}

As the second extension to distill off aspects other than style, we use both {\it nearby} and {\it all} contexts ($\C{near}$ and $\C{all}$).
As Figure~\ref{fig:method} shows, both the vector $\VEC{v}_{w}$ and $\tilde{\VEC{v}}_w$ of each word $w\in\mathcal V$ are divided into two vectors:
\begin{align}
   \VEC{v}_w = \VEC{x}_w \oplus \VEC{y}_w,\;\;
   \tilde{\VEC{v}}_w = \tilde{\VEC{x}}_w \oplus \tilde{\VEC{y}}_w
   \text{,}
\end{align}
where $\oplus$ denotes vector concatenation.
Vectors $\VEC{x}_{w}$ and $\tilde{\VEC{x}}_w$ indicate the style-sensitive part of $\VEC{v}_w$ and $\tilde{\VEC{v}}_w$ respectively.
Vectors $\VEC{y}_w$ and $\tilde{\VEC{y}}_w$ indicate the syntactic/semantic-sensitive part of $\VEC{v}_w$ and $\tilde{\VEC{v}}_w$ respectively.
For training,
when the context words are near the target word ($\C{near}$), we update both the style-sensitive vectors ($\tilde{\VEC{x}}_{w_t}$, $\VEC{x}_c$) and the syntactic/semantic-sensitive vectors ($\tilde{\VEC{y}}_{w_t}$, $\VEC{y}_c$), i.e., $\tilde{\VEC{v}}_{w_t}$, $\VEC{v}_c$.
Conversely, when the context words are far from the target word ($\C{dist}$), we only update the style-sensitive vectors ($\tilde{\VEC{x}}_{w_t}$, $\VEC{x}_c$).
Formally, the prediction probability is calculated as follows:
\begin{align}
P_1^{}(w_{t}|\C{near}) &\propto \exp\biggl(\!\tilde{\VEC{v}}_{w_t} \cdot \Cave{near}\!\biggr)
\text{,}
\\
P_2^{}(w_{t}|\C{dist}) &\propto \exp\biggl(\!\tilde{\VEC{x}}_{w_t} \cdot \CaveS{dist}\!\biggr)
\text{.}
\end{align}
At the time of learning, two prediction probabilities (loss functions) are alternately computed, and the word vectors are updated.
We refer to this method using the two-fold contexts separately as the \modelname{CBOW-sep-ctx}.

\section{Experiments}
We investigated which word vectors capture the stylistic, syntactic, and semantic similarities.

\subsection{Settings}
\paragraph{Training and Test Corpus}
We collected Japanese fictional stories from the Web to construct the dataset.
The dataset contains approximately $30$M utterances of fictional characters.
We separated the data into a $99$\%--$1$\% split for training and testing.
In Japanese, the function words at the end of the sentence often exhibit style (e.g., {\it desu}+{\it wa}, {\it desu}+{\it ze}\footnote{These words mean the verb {\it be} in English.};) therefore, we used an existing lexicon of multi-word functional expressions~\cite{miyazaki_15}.
Overall, the vocabulary size $\abs{\mathcal V}$ was $100$K.

\paragraph{Hyperparameters}
We chose the dimensions of both the style-sensitive and the syntactic/semantic-sensitive vectors to be $300$, and the dimensions of the baseline CBOWs were $300$.  
The learning rate was adjusted individually for each part in
$\{\VEC{x}_w, \VEC{y}_w, \tilde{\VEC{x}}_w, \tilde{\VEC{y}}_w\}$
such that ``the product of the learning rate and the expectation of the number of updates'' was a fixed constant.
We ran the optimizer with its default settings from the implementation of \citet{mikolov_13iclr}.
The training stopped after 10 epochs.
We fixed the nearby window width to $\delta=5$.

\subsection{Stylistic Similarity Evaluation}
\subsubsection{Data Construction}
To verify that our models capture the stylistic similarity, we evaluated our style-sensitive vector $\VEC{x}_{w_t}$ by comparing to other word vectors on a novel artificial task matching human stylistic similarity judgments.
For this evaluation, we constructed a novel dataset with human judgments on the stylistic similarity between word pairs by performing the following two steps.
First, we collected only style-sensitive words from the test corpus because some words are strongly associated with stylistic aspects~\cite{kinsui2003vaacharu,teshigawara2011modern} and, therefore, annotating random words for stylistic similarity is inefficient.
We asked crowdsourced workers to select style-sensitive words in utterances.
Specifically, for the crowdsourced task of picking “style-sensitive” words, we provided workers with a word-segmented utterance and asked them to pick words that they expected to be altered within different situational contexts (e.g., characters, moods, purposes, and the background cultures of the speaker and listener.).
Then, we randomly sampled $1,000$ word pairs from the selected words and asked $15$ workers to rate each of the pairs on five scales (from $-2$: ``{\it The style of the pair is different}'' to $+2$: ``{\it The style of the pair is similar}''), inspired by the syntactic/semantic similarity dataset~\cite{lev_02,emnlp16simverb}.
Finally, we picked only word pairs featuring clear worker agreement in which more than $10$ annotators rated the pair with the same sign,
which consisted of random pairs of highly agreeing style-sensitive words.
Consequently, we obtained $399$ word pairs with similarity scores.
To our knowledge, this is the first study that created an evaluation dataset to measure the lexical stylistic similarity.

In the task of selecting style-sensitive words,
the pairwise inter-annotator agreement was moderate (Cohen's kappa $\kappa$ is $0.51$).
In the rating task,
the pairwise inter-annotator agreement for two classes ($\{-2, -1\}$ or $\{+1, +2\}$) was fair (Cohen's kappa $\kappa$ is $0.23$).
These statistics suggest that, at least in Japanese, native speakers share a sense of style-sensitivity of words and stylistic similarity between style-sensitive words.

\subsubsection{Stylistic Sensitivity}
We used this evaluation dataset to compute the Spearman rank correlation ($\rho_{style}$) between the cosine similarity scores between the learned word vectors $\cos(\VEC{v}_{w}, \VEC{v}_{w'})$ and the human judgements.
Table~\ref{table:result} shows the results on its left side.
First, our proposed model, \modelname{CBOW-all-ctx} outperformed the baseline \modelname{CBOW-near-ctx}.
Furthermore, the $\VEC{x}$ of \modelname{CBOW-dist-ctx} and \modelname{CBOW-sep-ctx} demonstrated better correlations for stylistic similarity judgments ($\rho_{style}=56.1$ and $51.3$, respectively).
Even though the $\VEC{x}$ of \modelname{CBOW-sep-ctx} was trained with the same context window as \modelname{CBOW-all-ctx},
the style-sensitivity was boosted by introducing joint training with the near context.
\modelname{CBOW-dist-ctx}, which uses only the distant context, slightly outperforms \modelname{CBOW-sep-ctx}.
These results indicate the effectiveness of training using a wider context window.

\begin{table}[t]
  \centering
  \scalebox{0.8}{
  \begin{tabular}{rcccc}
    \hline
    \multicolumn{1}{c|}{\multirow{2}{*}{Model}}             & \multicolumn{1}{c}{\multirow{2}{*}{$\rho_{style}$}} & \multicolumn{1}{c}{\multirow{2}{*}{$\rho_{sem}$}} & \multicolumn{2}{c}{\textsc{SyntaxAcc}}\\
    \multicolumn{1}{l|}{}                                           & \multicolumn{1}{c}{}                          & & \multicolumn{1}{c}{@5} & \multicolumn{1}{c}{@10} \\ \hline
    \multicolumn{1}{l|}{\modelname{CBOW-near-ctx}}                  & 12.1            &27.8 &86.3 & 85.2  \\
    \multicolumn{1}{l|}{\modelname{CBOW-all-ctx}}                    & 36.6            & 24.0 &85.3 &  84.1 \\ \hline
    \multicolumn{1}{l|}{\modelname{CBOW-dist-ctx}}                  & \textbf{56.1} & 15.9 &59.4 & 58.8  \\ \hline
    \multicolumn{1}{l|}{\modelname{CBOW-sep-ctx}}                  &  &  &  & \\
     \multicolumn{1}{l|}{\ \ $\VEC{x}$ (Stylistic)}                   & \textbf{51.3} & \textbf{28.9} &68.3 & 66.2  \\
     \multicolumn{1}{l|}{\ \ $\VEC{y}$ (Syntactic/semantic)} & 9.6              &18.1 &\textbf{88.0} & \textbf{87.0}  \\ \hline
  \end{tabular}
  }\\
   \caption{Results of the quantitative evaluations.}
  \label{table:result}
\end{table}

\begin{table*}[t]
  \centering
  \footnotesize
  \begin{tabular}{r|l|l|l}
  \hline
  \multicolumn{2}{c|}{\multirow{2}{*}{Word $w$}}&\multicolumn{2}{c}{The top similar words $\{w'\}$ to $w$ w.r.t. cosine similarity}\\ \cline{3-4}
  \multicolumn{2}{c|}{}
  &$\cos(\VEC{x}_w, \VEC{x}_{w'})$ (stylistic half)
  &$\cos(\VEC{y}_w, \VEC{y}_{w'})$ (syntactic/semantic half)
  \\ \hline \hline
  \parbox[t]{2mm}{\multirow{14}{*}{\rotatebox[origin=c]{90}{Japanese}}}
  &\ja{俺} (I; male, colloquial)
  &\ja{おまえ} (you; colloquial, rough),
  &\ja{僕} (I; male, colloquial, childish),
  \\
  &
  &\ja{あいつ} (he/she; colloquial, rough),
  &\ja{あたし} (I; female, childish),
  \\
  &
  &\ja{ねーよ} (not; colloquial, rough, male)
  &\ja{私} (I; formal)
  \\
  \cline{2-4}
  &\ja{拙者} (I; classical${}^{*}$)
  &\ja{でござる}(be; classical),
  &\ja{僕} (I; male, childish),
  \\
  &\hspace{.5em}${}^{*}$\! e.g., samurai, ninja
  &\ja{ござる}(be; classical),
  &\ja{俺} (I; male, colloquial),
  \\
  &
  &\ja{ござるよ}(be; classical)
  &\ja{私} (I; formal)
  \\
  \cline{2-4}
  &\ja{かしら} (wonder; female)
  &\ja{わね} (QUESTION; female),
  &\ja{かな} (wonder; childish),
  \\
  &
  &\ja{ないわね} (not; female),
  &\ja{でしょうか} (wonder; fomal),
  \\
  &
  &\ja{わ} (SENTENCE-FINAL; female)
  &\ja{かしらね} (wonder; female)
  \\
  \cline{2-4}
  &\ja{サンタ} (Santa Clause; shortened)
  &\ja{サンタクロース} (Santa Clause; -),
  &\ja{お客} (customer; little polite),
  \\
  &
  &\ja{トナカイ} (reindeer; -),
  &\ja{プロデューサー} (producer; -),
  \\
  &
  &\ja{クリスマス} (Christmas; -)
  &\ja{メイド} (maid; shortened)
  \\
  \hline\hline
  \parbox[t]{2mm}{\multirow{4}{*}{\rotatebox[origin=c]{90}{English}}}
  &shit
  &fuckin, fuck, goddamn
  &shitty, crappy, sucky\\ \cline{2-4}
  &hi
  &hello, bye, hiya, meet
  &goodbye, goodnight, good-bye\\ \cline{2-4}
  &guys
  &stuff, guy, bunch
  &boys, humans, girls\\ \cline{2-4}
  &ninja
  &shinobi, genin, konoha
  &shinobi, pirate, soldier\\
  \hline
  \end{tabular}
   \caption{The top similar words for the style-sensitive and syntactic/semantic vectors learned with proposed model, \modelname{CBOW-sep-ctx}. Japanese words are translated into English by the authors. Legend: (translation; impression).}
     \label{table:result_topn}
\end{table*}

\subsection{Syntactic and Semantic Evaluation}
We further investigated the properties of each model using the following criterion:
(1) the model's ability to capture the syntactic aspect was assessed through a task predicting part of speech (POS) and
(2) the model's ability to capture the semantic aspect was assessed through a task calculating the correlation with human judgments for semantic similarity.


\subsubsection{Syntactic Sensitivity}

First, we tested the ability to capture syntactic similarity of each model by checking whether the POS of each word was the same as the POS of a neighboring word in the vector space.
Specifically, we calculated \textsc{SyntaxAcc}@$N$ defined as follows:
\begin{align}
   \frac{1}{\abs{\mathcal V} N}\sum_{w\in \mathcal V}\sum_{\,w'\in \mathcal{N}(w)} \hspace{-4pt}\mathbb{I}[\mathrm{POS}(w) \!=\! \mathrm{POS}(w')]
   \text{,}\!
\end{align}
where $\mathbb{I}[\text{condition}] = 1$ if the condition is true and $\mathbb{I}[\text{conditon}] = 0$ otherwise,
the function $\mathrm{POS}(w)$ returns the actual POS tag of the word $w$,
and $\mathcal{N}(w)$ denotes the set of the $N$ top similar words $\{w'\}$ to $w$ w.r.t. $\cos(\VEC{v}_w,\VEC{v}_{w'})$ in each vector space.

Table~\ref{table:result} shows \textsc{SyntaxAcc}@$N$ with $N = 5$ and $10$.
For both $N$, the $\VEC{y}$ (the syntactic/semantic part) of \modelname{CBOW-near-ctx}, \modelname{CBOW-all-ctx} and \modelname{CBOW-sep-ctx} achieved similarly good.
Interestingly, even though the $\VEC{x}$ of \modelname{CBOW-sep-ctx} used the same context as that of \modelname{CBOW-all-ctx},
the syntactic sensitivity of $\VEC{x}$ was suppressed.
We speculate that the syntactic sensitivity was distilled off by the other part of the \modelname{CBOW-sep-ctx} vector, i.e., $\VEC{y}$ learned using only the \textit{near} context, which captured more syntactic information.
In the next section, we analyze \modelname{CBOW-sep-ctx} for the different characteristics of $\VEC{x}$ and $\VEC{y}$.


\subsubsection{Semantic and Topical Sensitivities}
To test the model's ability to capture the semantic similarity, we also measured correlations with the Japanese Word Similarity Dataset (JWSD)~\cite{sakaizawa18}, which consists of $4,\!000$  Japanese word pairs annotated with semantic similarity scores by human workers.
For each model, we calculate and show  the Spearman rank correlation score ($\rho_{sem}$) between the cosine similarity score $\cos(\VEC{v}_w, \VEC{v}_{w'})$ and the human judgements on JWSD in Table~\ref{table:result}\footnote{Note that the low performance of our baseline ($\rho_{sem}\!=\!27.8$ for \modelname{CBOW-near-ctx}) is unsurprising comparing to English baselines (cf., \citet{taguchi17}).}.
\modelname{CBOW-dist-ctx} has the lowest score ($\rho_{sem}\!=\!15.9$); however, surprisingly, the stylistic vector $\VEC{x}_{w_t}$ has the highest score ($\rho_{sem}\!=\!28.9$), while both vectors have a high $\rho_{style}$.
This result indicates that the proposed stylistic vector $\VEC{x}_{w_t}$ captures not only the stylistic similarity but also the captures semantic similarity, contrary to our expectations (ideally, we want the stylistic vector to capture only the stylistic similarity).
We speculate that this is because not only the \emph{style} but also the \emph{topic} is often consistent in single utterances.
For example, ``\ja{サンタ} (Santa Clause)'' and ``\ja{トナカイ} (reindeer)'' are topically relevant words and these words tend to appear in a single utterance.
Therefore, stylistic vectors $\{\VEC{x}_{w}\}$ using all the context words in an utterance also capture the topic relatedness.
In addition, JWSD contains topic-related word pairs and synonym pairs; therefore the word vectors that capture the topic similarity have higher $\rho_{sem}$.
We will discuss this point in the next section.

\subsection{Analysis of Trained Word Vectors}
Finally, to further understand what types of features our \modelname{CBOW-sep-ctx} model acquired,
we show some words\footnote{We arbitrarily selected style-sensitive words from our stylistic similarity evaluation dataset.} with the four most similar words in Table~\ref{table:result_topn}.
Here, for English readers, we also report a result for English\footnote{We trained another \modelname{CBOW-sep-ctx} model on an English fan-fiction dataset that was collected from the Web\\ (\url{https://www.fanfiction.net/}).}.
The English result also shows an example of the performance of our model on another language.
The left side of Table~\ref{table:result_topn} (for stylistic vector $\VEC{x}$) shows the results.
We found that the Japanese word ``\ja{拙者} (I; classical)''  is similar to ``\ja{ござる} (be; classical)'' or words containing it (the second row of Table~\ref{table:result_topn}).
The result looks reasonable, because words such as ``\ja{拙者} (I; classical)'' and  ``\ja{ござる} (be; classical)''  are typically used by Japanese {\it Samurai} or {\it Ninja}.
We can see that the vectors captured the similarity of these words, which are stylistically consistent across syntactic and semantic varieties.
Conversely, the right side of the table (for the syntactic/semantic vector $\VEC{y}$) shows that the word ``\ja{拙者} (I; classical)'' is similar to the personal pronoun (e.g., ``\ja{僕} (I; male, childish)'').
We further confirmed that $15$ the top similar words are also personal pronouns (even though they are not shown due to space limitations).
These results indicate that the proposed \modelname{CBOW-sep-ctx} model jointly learns two different types of lexical similarities, i.e., the stylistic and syntactic/semantic similarities in the different parts of the vectors.
However, our stylistic vector also captured the topic similarity, such as ``\ja{サンタ} (Santa Clause)'' and ``\ja{トナカイ} (reindeer)'' (the fourth row of Table~\ref{table:result_topn}).
Therefore, there  is still room for improvement in capturing the stylistic similarity.

\section{Conclusions and Future Work}
This paper presented the unsupervised learning of style-sensitive word vectors, which extends CBOW by distinguishing nearby contexts and wider contexts.
We created a novel dataset for style, where the stylistic similarity between word pairs was scored by human.
Our experiment demonstrated that our method leads word vectors to distinguish the stylistic aspect and other semantic or syntactic aspects.
In addition, we also found that our training cannot help confusing some styles and topics.
A future direction will be to addressing the issue by further introducing another context such as a document or dialog-level context windows, where the topics are often consistent but the styles are not.

\section*{Acknowledgments}
This work was supported by JSPS KAKENHI Grant Number 15H01702.
We thank our anonymous reviewers for their helpful comments and suggestions.

\bibliography{acl2018}
\bibliographystyle{acl_natbib}


\end{document}